\documentclass[10pt,letterpaper]{article}
\usepackage[top=0.85in,left=1.0in,footskip=0.75in]{geometry}

\usepackage{changepage}

\usepackage[utf8]{inputenc}

\usepackage{textcomp,marvosym}

\usepackage{fixltx2e}

\usepackage{amsmath,amssymb}

\usepackage{cite}

\usepackage{nameref,hyperref}

\usepackage[right]{lineno}

\usepackage{microtype}
\DisableLigatures[f]{encoding = *, family = * }

\usepackage{rotating}

\usepackage{color}

\raggedright
\usepackage[aboveskip=1pt,labelfont=bf,labelsep=period,justification=raggedright,singlelinecheck=off]{caption}

\makeatletter
\renewcommand{\@biblabel}[1]{\quad#1.}
\makeatother

\date{}

\newcommand{\acc}{$\mathit{Acc}$}
\newcommand{\accone}{$\mathit{Acc}\!\pm\!1$}
\newcommand{\favg}{$\overline{F_{1}}$}
\newcommand{\alfa}{$\mathit{Alpha}$}
\newcommand{\kaint}{\mathit{{Alpha}_{int}}}
\newcommand{\kanom}{\mathit{{Alpha}_{nom}}}

\newcommand{\sn}{\hphantom{0}}

\usepackage{color}

\usepackage{verbatim}

\begin{document}

\begin{flushleft}
{\Large
\textbf\newline{Multilingual Twitter Sentiment Classification:\\[1ex]
The Role of Human Annotators}
}
\newline
\\
Igor Mozeti\v{c}*,
Miha Gr\v{c}ar,
Jasmina Smailovi\'{c}*
\\
\bigskip
Department of Knowledge Technologies, Jo\v{z}ef Stefan Institute, Ljubljana, Slovenia
\\
\bigskip

%
%






* Igor.Mozetic@ijs.si (IM); Jasmina.Smailovic@ijs.si (JS)

\end{flushleft}
\section*{Abstract}
What are the limits of automated Twitter sentiment classification?
We analyze a large set of manually labeled tweets in different languages,
use them as training data, and construct automated classification models.
It turns out that the quality of classification models depends much more
on the quality and size of training data than on the type of the model trained.
Experimental results indicate that there is no statistically significant difference
between the performance of the top classification models.
We quantify the quality of training data by applying various annotator
agreement measures, and identify the weakest points of different datasets.
We show that the model performance approaches the inter-annotator 
agreement when the size of the training set is sufficiently large.
However, it is crucial to regularly monitor the self- and inter-annotator agreements
since this improves the training datasets and consequently the model performance.
Finally, we show that there is strong evidence that humans perceive the
sentiment classes (negative, neutral, and positive) as ordered.


\section{Introduction}

Sentiment analysis is a form of shallow semantic analysis of texts.
Its goal is to extract opinions, emotions or attitudes towards 
different objects of interest \cite{liu2012sentiment,liu2015book}.
For example, one might be interested in consumers opinion about
products, voters attitude towards political parties, or investors
expectations about stocks.
From the first approaches in 2000s, sentiment analysis gained considerable
attention with massive growth of the web and social media.
Different forms of textual information are becoming easily accessible 
(e.g., news, blogs, reviews, Facebook comments, Twitter posts, etc.),
and different approaches to sentiment analysis were developed.

There are two prevailing approaches to large-scale sentiment analysis:
(i) lexicon-based and (ii) machine learning.
In the first case, the sentiment in the text is computed from the set of
sentiment-bearing words identified in the text.
In the second case, a sentiment classification model is constructed first,
from a large set of sentiment labeled texts, 
and then applied to the stream of unlabelled texts.
The model has the form of a function that maps features extracted from the
text into sentiment labels (which typically have discrete values:
negative, neutral, or positive).
In both approaches, one needs a considerable involvement of humans, at least
initially. Humans have to label their perception of the sentiment expressed
either in individual words or in short texts.
This sentiment labeling is language-, domain- and often even topic-specific.

An example of a lexicon-based approach that involves a massive human 
sentiment labeling of words is described by Dodds et al. \cite{Dodds2015lang}.
They collected around 5 million human sentiment assessments of 10,000 
common words, each in 10 languages and labeled 50 times.
Another well-known sentiment lexicon is
SentiWordNet \cite{baccianella2010sentiwordnet}, constructed semi-automatically 
for over 100,000 words, but limited to English only.

In this paper we analyze a set of over 1.6 million Twitter posts,
in 13 European languages, labeled for sentiment by human annotators.
The labeled tweets are used as training data to train sentiment classifiers
for different languages. 
An overview of the state-of-the-art of Twitter sentiment analysis 
is given in \cite{martinez2014sentiment}.
A more recent overview of the lexicon-based and machine learning methods,
and their combination, is in \cite{kolchyna2015twitter}.

We focus on the quantity and quality of the labeled
tweets, and their impact on the performance of sentiment classifiers.
The quality of the labeled tweets is estimated from 
the agreement between human annotators. 
The main hypothesis of the paper is that the annotators agreement 
provides an upper bound for the classifier performance. 

There are several more specific research questions we address: \\
(1) Are the sentiment classes ordered? \\
(2) Which evaluation measures are appropriate to quantify and compare
the labeled data quality and classifiers performance? \\
(3) How to estimate the quality of the training data? \\
(4) How to compare and select appropriate classifiers? \\
(5) What are acceptable levels of the annotators agreement? \\
(6) How many labeled Twitter posts are needed for training a sentiment classifier? \\

In the paper we present three lines of experiments and results.
One is related to manual annotation of Twitter posts and estimations of
their quality and dataset properties.
Another is about training sentiment classifiers, their performance
and comparisons.
The third line compares the labeled data quality with the classifier
performance and provides support for our main hypothesis.

The paper is organized as follows.
In the Results and Discussion section we provide the main results on
the comparison of the annotators agreement and classifiers performance.
We briefly outline the main evaluation measure used and the datasets analyzed.
The evaluation procedures and methods are just sketched, to facilitate
the discussion of the results---all the details are in the Methods section.
The main emphasis is on an in-depth analysis of the datasets.
We consider their evolution through time, as new tweets get annotated,
and how the performance of the classifiers varies with time.
We also discuss the effects of different distributions of the
training and application datasets.

Conclusions provide answers to the research questions addressed,
and give short- and long-term directions of future research.

The Methods section provides all the details about the first two
lines of experiments and results, specifically about the data,
annotations, and sentiment classifiers.
We define four evaluation measures, common in the fields of
inter-rater agreement and machine learning. 
The measures are used to compute the self- and inter-annotator
agreements for all the datasets. 
From these results we derive evidence that human annotators perceive the 
sentiment classes as ordered.
We present the related work on methods used for the Twitter sentiment
classification, and publicly available labeled datasets.
We compare the performance of six selected classifiers by applying a
standard statistical test.
We give the necessary details of the evaluation procedure and the 
standard Twitter pre-processing steps.

In the following subsection we give an overview of the related work
on automated sentiment classification of Twitter posts.
We summarize the published labeled sets used for training the 
classification models, and the machine learning
methods applied for training.
Most of the related work is limited to English texts only.

\paragraph{Contributions.}
We provide a large corpus of sentiment labeled tweets, in different
languages and of varying quality. 
The collected set of over 1.6 million manually labeled tweets
is by far the largest dataset reported in the literature, and we make it
publicly available.
We expect the corpus to be a fruitful and
realistic test-bed for various classification algorithms.
We apply four evaluation measures and show that two of them
are more appropriate to evaluate sentiment classifiers.
Additionally, the same measures are used to evaluate the quality of 
training data, thus providing the means to monitor the annotation process.
We do not address various options in Twitter
pre-processing and feature selection.
Instead, we use the same standard parameter settings to get an unbiased
comparison of six sentiment classifiers.

\section{Results and Discussion}

In this paper we analyze a large set of sentiment labeled tweets.
We assume a sentiment label takes one of three possible values:
\textit{negative}, \textit{neutral}, or \textit{positive}.
The analysis sheds light on two aspects of the data: the quality of human
labeling of the tweets, and the performance of the sentiment classification
models constructed from the same data.
The main idea behind this analysis is to use the same
evaluation measures to estimate both, the quality of human annotations
and the quality of classification models.
We argue that the performance of a classification model 
is primarily limited by the quality of the labeled data.
This, in turn, can be estimated by the agreement between the human annotators.

\paragraph{Evaluation measures.}
The researchers in the fields of inter-rater agreement and machine learning
typically employ different evaluation measures.
We report all the results in terms of four selected measures
which we deem appropriate for the three-valued sentiment 
classification task (the details are in the Evaluation measures subsection in Methods).
In this section, however, the results are summarized only in terms of
Krippendorff's Alpha-reliability (\alfa) \cite{Krippendorff2012},
to highlight the main conclusions.
\alfa\, is a generalization of several specialized agreement measures.
When annotators agree perfectly or when a model perfectly classifies the data,
\alfa\;$=1$. When the level of agreement equals the agreement by chance, \alfa\;$=0$. 
There are several instances of \alfa. All the results are reported
here are in terms of $\kaint$ (interval)
which takes into account the ordering of sentiment values
and assigns higher penalty to more extreme disagreements.
The justification for this choice is in the
subsection on Ordering of sentiment values in Methods.

\paragraph{Datasets.}
We analyze two corpora of data. The first consists of 13 language datasets,
with over 1.6 million annotated tweets, by far the largest sentiment corpus
made publicly available. The languages covered are:
\textbf{Albanian, Bulgarian, English, German, Hungarian, Polish, Portuguese,
Russian, Ser/Cro/Bos} (a joint set of Serbian, Croatian, and Bosnian tweets,
the languages difficult to distinguish on Twitter), 
\textbf{Slovak, Slovenian, Spanish}, and \textbf{Swedish}. 

The second corpus of data comes from four applications of sentiment 
classification, which we already published.
These tweets are domain-specific and provide novel insights and
lessons learned when analyzed with the same methods as the language datasets.
The application datasets are:
\textbf{Facebook(it)} - the Facebook comments on conspiracy theories in Italian,
to study the emotional dynamics \cite{Zollo2015facebook},
\textbf{DJIA30} - tweets about the Dow Jones stocks, 
to analyze the effects of Twitter sentiment on their price movements
\cite{Ranco2015eventstudy},
\textbf{Environment} - tweets about environmental issues, to compare the 
sentiment leaning of different communities \cite{Sluban2015sentlean}, and
\textbf{Emojis} - a subset of the tweets from the above 13 language datasets which 
contain emojis, used to derive the emoji sentiment lexicon \cite{Kralj2015emojis}.
The details about the datasets in terms of their size, sentiment distribution,
and the time of the posts are in the Datasets subsection in Methods.

\subsection{The limits of performance}

Determining sentiment expressed in a tweet is not an easy task,
and depends on subjective judgment of human annotators.
Annotators often disagree between themselves, and even an individual
is not always consistent with her/himself.
There are several reasons for disagreements, such as:
inherent difficulty of the task (e.g., estimating the ``sentiment''
about the future stock movement),
different vocabularies used in different domains (e.g., financial
markets vs. environmental issues),
topic drift in time (e.g., events which abruptly shift the topic
of discussions on Twitter), or
simply a poor quality of the annotator's work.
In the data we analyze, we observe all the above issues, try
to identify them by computational means, and draw lessons how
the annotation process should be conducted in the future.

\paragraph{Annotator agreements.}
During the manual sentiment labeling of tweets, a fraction of tweets
(about 15\%) was intentionally duplicated to be annotated twice,
either by the same annotator or by two different annotators
(see details in the Datasets subsection in Methods).
From multiple annotations of the same annotator we compute
the \textbf{self-agreement}, and from multiple annotations
by different annotators we compute the 
\textbf{inter-annotator agreement} (abbreviated as inter-agreement). 
The confidence intervals for the agreements are estimated by 
bootstrapping \cite{Mooney1993}. The detailed results are in
the Annotator agreements subsection in Methods.
It turns out that the self-agreement is a good measure to
identify low quality annotators, and that the inter-annotator agreement
provides a good estimate of the objective difficulty of the task,
unless it is too low.

\paragraph{Model evaluation.}
To manually label over 1.6 million tweets requires a considerable effort.
The purpose of this effort is to use the labeled data to built sentiment 
classification models for each of the 13 languages.
A classification model can then be applied to unlabeled data in
various application scenarios, as was the case with our four
application datasets.

A classification model can be build by any suitable supervised machine
learning method. To evaluate the model, a standard approach in machine
learning is to use 10-fold cross-validation. The whole labeled set
is partitioned into 10 folds, one is set apart for testing,
and the remaining nine are used to train the model and evaluate it
on the test fold. The process is repeated 10 times until each fold
is used for testing exactly once. The reported evaluation results are
the average of 10 tests, and the confidence intervals are
estimated from standard deviations.

We constructed and evaluated six different classification models 
for each labeled language dataset.
The results for the application datasets
are extracted from the original papers.
Our classifiers are all based on Support Vector Machines (SVM) \cite{Vapnik1995},  
and for reference we also constructed a Naive Bayes classifier \cite{Russell2003}.
Detailed results are in the Classification models performance subsection in Methods.
When comparing the classifiers' performance with the Friedman-Nemenyi
test \cite{Friedman1940,Nemenyi1963}, it turns out that there is 
no statistically significant difference between most of them
(see the Friedman-Nemenyi test subsection in Methods).
For subsequent analyses and comparisons, we selected
the TwoPlaneSVMbin classifier that is always in the group of top classifiers
according to two most relevant evaluation measures.

\begin{figure}[!h]
\begin{center}
\includegraphics[width=\textwidth]{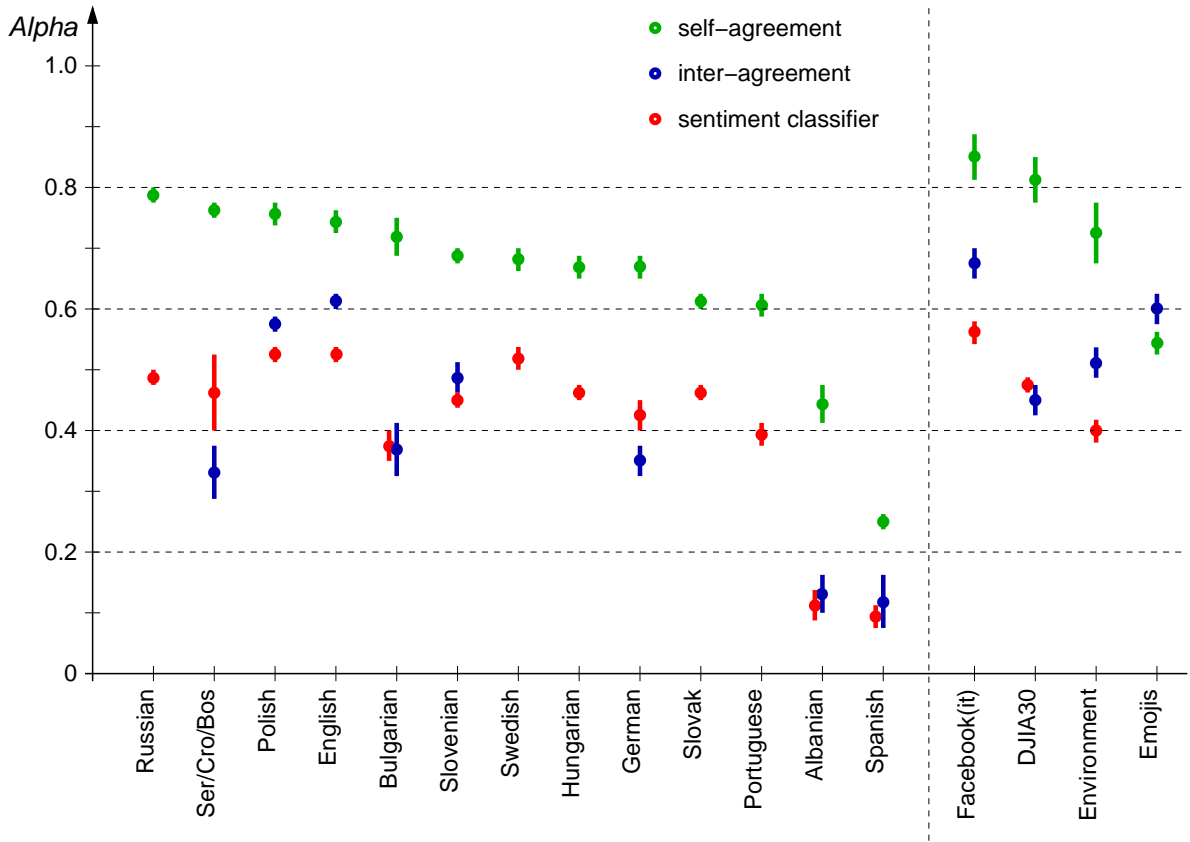}
\caption{\textbf{Comparison of annotators self-agreement (green), 
the inter-annotator agreement (blue), and an automated sentiment classifier 
(TwoPlaneSVMbin, red) in terms of Krippendorff's \alfa.}
On the left-hand side are the 13 language datasets, and on the right-hand side the four
application datasets. The datasets are ordered by decreasing self-agreement.
The error bars indicate estimated 95\% confidence intervals.} 
\label{fig:agree}
\end{center}
\end{figure}

\paragraph{Comparative analyses.}
The main results of this paper are summarized in Fig~\ref{fig:agree}.
It shows a comparison of the self-agreement, the inter-annotator agreement,
and the TwoPlaneSVMbin classifier performance, for the 13 language
datasets and the four application datasets.

The \textbf{self-agreement} for most of the datasets is above 0.6.
The exceptions, Albanian and Spanish, indicate low quality annotators
which should be eliminated from further considerations.
In the applications corpus,
the Emojis dataset is the only one with the self-agreement lower than
the inter-annotator agreement, due to a high number of low quality 
Spanish annotations included.
The other three application datasets have
relatively high self-agreement (0.7--0.9, due to more carefully 
selected annotators), and higher variability (due to a lower number
of tweets annotated twice, 2--4\% only).

The \textbf{inter-annotator agreement} varies a lot, and is always
lower than the self-agreement, except for Emojis.
The high inter-annotator agreement for Facebook(it) is consistent with
the high self-agreement.
Values below 0.2 (Albanian and Spanish) indicate low quality
annotators, consistent with the low self-agreement.
Values in the range between 0.3--0.4 (Ser/Cro/Bos, Bulgarian, and German)
indicate a problem with the annotation process,
and are discussed in more detail in the next subsection. 

The \textbf{classifier performance} is typically in the range
between 0.4--0.6. Notable exceptions are Albanian and Spanish,
with the performance barely above random, but very close to
the inter-annotator agreement. More interesting are the datasets with
a relatively low performance, around 0.4, that cannot be explained
by low quality annotations alone: Ser/Cro/Bos, Bulgarian, German, 
Portuguese, and Environment. They are analyzed in the next subsections. 

The main hypothesis of this paper is that the inter-annotator agreement 
approximates an upper bound for a classifier performance. 
In Fig~\ref{fig:agree} we observe three such cases where the classifier
performance, in the range 0.4--0.6, 
approaches its limit: Polish, Slovenian, and DJIA30.
There are also three cases where there still appears a gap between
the classifier performance and the inter-annotator agreement:
English, Facebook(it), and Environment.
In order to confirm the hypothesis, we analyze the evolution of
the classifiers performance through time and check if the performance
is still improving or was the plateau already reached.
This is not always possible:
There are datasets where only one annotator was engaged and
for which there is no inter-annotator agreement
(Russian, Swedish, Hungarian, Slovak, and Portuguese).
For them we can only draw analogies with the multiply annotated
datasets and speculate about the conclusions.

In the next two subsection we first analyze the language datasets,
and then the four application datasets.

\subsection{Language datasets analyses}

To label the 1.6 million tweets in the 13 languages, 83 native speakers were engaged,
and an estimated effort of 38 person-months was spent.
Can one reduce the efforts and focus them on more problematic datasets instead?
It seems, for example, that the annotation of over 200,000 Polish tweets
was an overkill. Worse, the annotation of over 250,000 Spanish tweets
was largely a waste of efforts, due to the poor annotation quality.

We perform a post-hoc analysis of the 13 language datasets by measuring
the performance of the sentiment classifiers through time. 
We emulate the evolution of the performance by feeding increasingly
large labeled sets into the classifier training process.
The labeled sets are ordered by the post time of the tweets,
so one can detect potential topic shifts during the Twitter discussions.
At each stage, the labeled set is increased by 10,000 tweets,
and the set accumulated so far is used for training and testing the classifier.
After each stage, the evaluation by 10-fold cross-validation is
performed and the results are reported in the following charts.
The final stage, when all the labeled sets are exhausted, 
corresponds to the results reported in Fig~\ref{fig:agree}.
In subsequent figures, the x-axis denotes labeled sets increases
by 10,000 tweets, the y-axis denotes the TwoPlaneSVMbin classifier
performance measured by \alfa, and the error bars are the 95\%
confidence intervals estimated from 10-fold cross-validations.
The inter-annotator agreement is represented by a blue 
line---it is constant and is computed from all the available data.

We identify five cases, characterized by different relations between
the classifier performance and the inter-annotator agreement:
(i) a performance gap still exists, (ii) a performance limit is approached, 
(iii) low inter-annotator agreement, (iv) topic shift, and 
(v) very low annotation quality.

\begin{figure}[!h]
\begin{center}
\includegraphics[width=\textwidth]{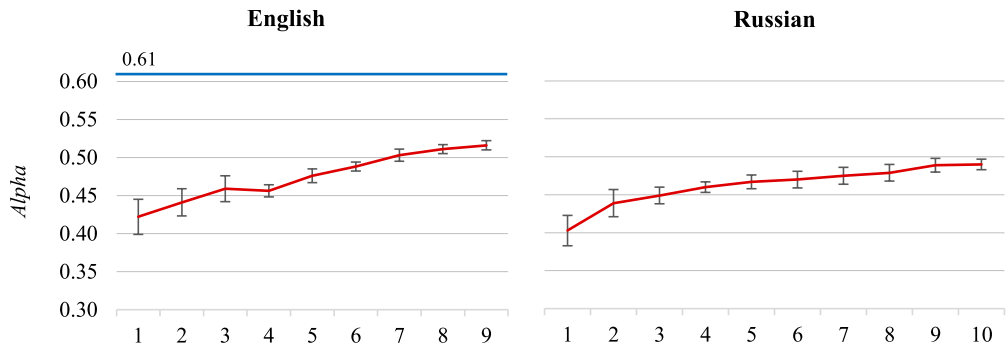}
\caption{\textbf{The English (left) and Russian (right) datasets.}
For English, there is still a gap (\alfa\, = 0.097) between the classifier (in red) 
and the inter-annotator agreement (in blue).}
\label{fig:eng-rus}
\end{center}
\end{figure}

\paragraph{Performance gap still exists.}
Fig~\ref{fig:eng-rus} (chart on the left) shows the evolution of the
\textbf{English} classifier performance, as it is fed increasingly large
training sets. On top (in blue) is the inter-annotator agreement line
(\alfa\, = 0.613). The classifier's \alfa\, is increasing from the
initial 0.422 to 0.516, but is still considerably below the inter-annotator
agreement. Despite the relatively large training set (around 90,000 
labeled tweets) there is still a performance gap and even
more annotations are needed to approach the inter-annotator agreement.

We observe a similar pattern with the \textbf{Russian} (Fig~\ref{fig:eng-rus}, 
chart on the right) and \textbf{Slovak} datasets (not shown).
The inter-annotator agreement
is unknown, but the classifier's performance is still increasing
from the initial \alfa\, of 0.403 to 0.490 for Russian, and
from the initial 0.408 to 0.460 for Slovak. The size of the labeled
sets for Russian is around 90,000, for Slovak around 60,000, and we
argue that more training data is needed to further improve the performance.

\begin{figure}[!h]
\begin{center}
\includegraphics[width=\textwidth]{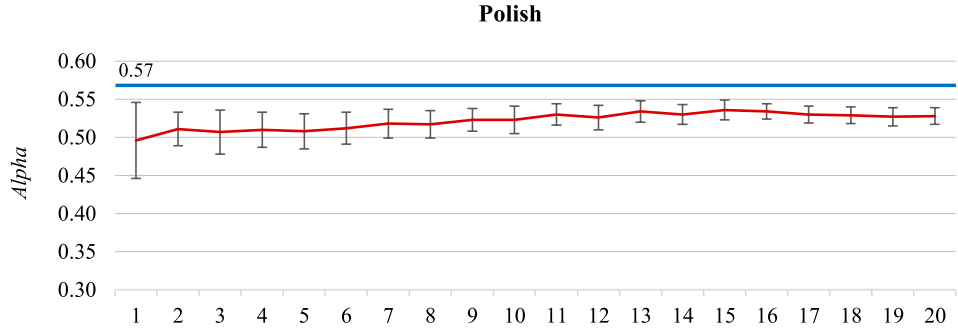}
\caption{\textbf{The Polish dataset.}
The classifier's peak performance (in red, \alfa\, = 0.536) is
at 150,000 labeled tweets.}
\label{fig:pol}
\end{center}
\end{figure}

\paragraph{Performance limit approached.}
A different pattern from the above can be observed in
Fig~\ref{fig:pol} for the \textbf{Polish} dataset.
After a slow improvement of the classifier's performance, the peak is
reached at around 150,000 labeled tweets, and afterwards the performance
remains stable and is even slightly decreasing.
The maximum \alfa\, is 0.536, close to the inter-annotator agreement of
0.571. At the same point, at 150,000 tweets, another performance measure,
\favg, also peaks at its maximum value, even above the corresponding
inter-annotator agreement.
These results suggest that beyond a certain point, when the classifier's
performance is ``close enough'' to the inter-annotator agreement,
it does not pay off to further label tweets by sentiment.
This is valid, however, only until a considerably new topic occurs.

\begin{figure}[!h]
\begin{center}
\includegraphics[width=\textwidth]{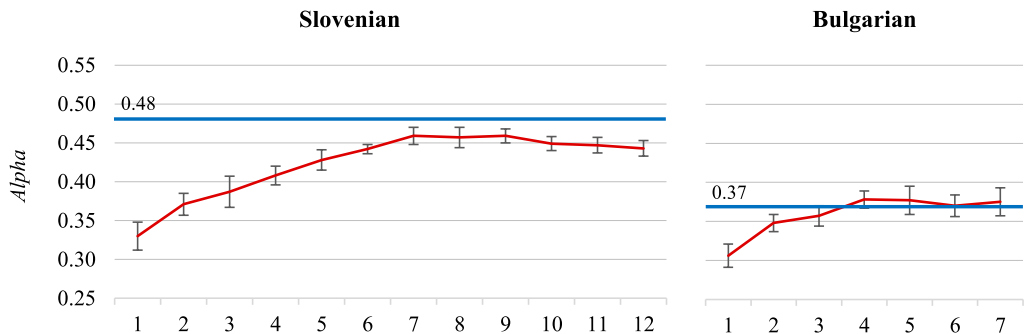}
\caption{\textbf{The Slovenian (left) and Bulgarian (right) datasets.}
The Slovenian classifier peak is at 70,000 tweets (\alfa\, = 0.459).
The Bulgarian classifier peak is at 40,000 tweets (\alfa\, = 0.378).}
\label{fig:slo-bul}
\end{center}
\end{figure}

Similar conclusions can be drawn for the \textbf{Slovenian} dataset
(Fig~\ref{fig:slo-bul}, chart on the left).
The classifier's performance reaches its peak earlier, at 70,000 tweets,
with the maximum \alfa\, of 0.459, as well as the maximum \favg.
\alfa\, is close to the inter-annotator agreement of 0.485,
and \favg\, even exceeds the corresponding agreement.
However, notice that the inter-annotator agreement for Slovenian
is almost 10\% points lower than for Polish.

We observe a similar pattern for the \textbf{Bulgarian} dataset 
(Fig~\ref{fig:slo-bul}, chart on the right).
The classifier's peak performance is reached even earlier, at 40,000 tweets
(\alfa\, is 0.378),
but the inter-annotator agreement
is also considerably lower, more than 10\% points below the Slovenian 
(\alfa\, is 0.367).
In such cases, when the inter-annotator agreement is ``too low''
(our estimate is when \alfa\, $< 0.4$), the inter-annotator agreement is a
poor estimator of the difficulty of the task, and should not be used
as a performance approximation. Instead, one could analyze the reasons for
the disagreements, as we do with cases in the following paragraphs.

\begin{figure}[!h]
\begin{center}
\includegraphics[width=\textwidth]{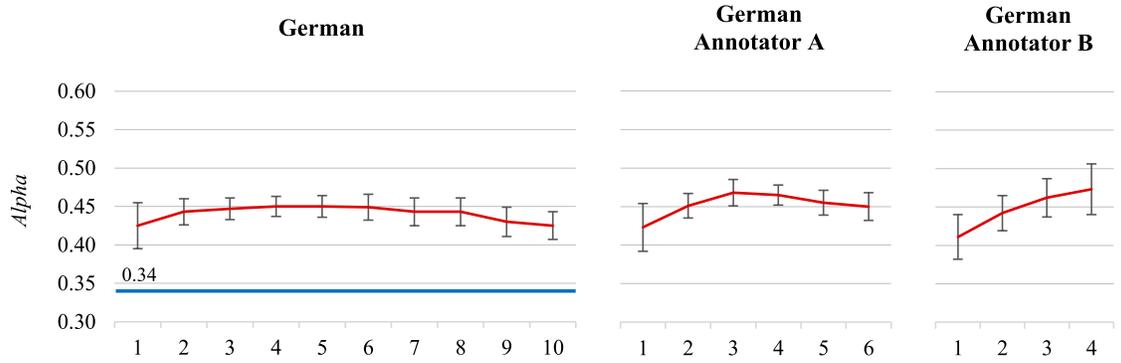}
\caption{\textbf{The German datasets.}
The complete dataset (left), and separate datasets labeled by the 
two main annotators (middle and right).}
\label{fig:ger}
\end{center}
\end{figure}

\paragraph{Low inter-annotator agreement.}
The inter-annotator agreement for the \textbf{German} dataset is low,
\alfa\, is 0.344.
The classifier's performance is higher already with the initial small datasets, 
and soon starts dropping (Fig~\ref{fig:ger}, chart on the left).
It turns out that over 90\% of the German tweets were labeled by two
annotators only, dubbed annotator A and B.
The annotation quality of the two annotators is very different,
the self-agreement \alfa\, for the annotator A is 0.590, 
and for the annotator B is 0.760.
We consider the German tweets labeled by A and B separately
(Fig~\ref{fig:ger}, charts in the middle and on the right).
The lower quality A dataset reaches its maximum at 30,000 tweets,
while the performance of the higher quality B dataset is still increasing.
There was also a relatively high disagreement between the two annotators
which resulted in a low classifier's performance.
A conclusions drawn from this dataset, as well as from the Bulgarian,
is that one should constantly monitor
the self- and inter-annotator agreements, and promptly notify the annotators 
as soon as the agreements drop too low.

\begin{figure}[!h]
\begin{center}
\includegraphics[width=\textwidth]{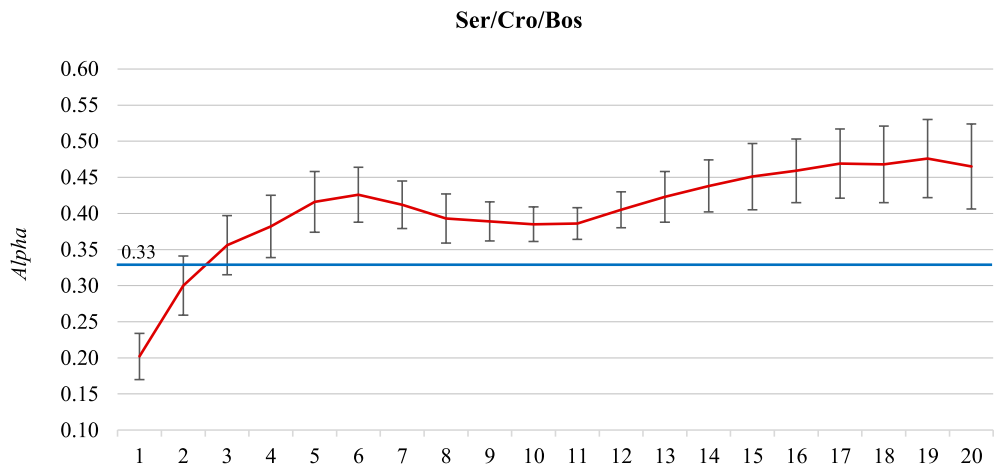}
\caption{\textbf{Joint Serbian/Croatian/Bosnian dataset.}
There is an oscillation in performance and high variability.}
\label{fig:scb}
\end{center}
\end{figure}

\newpage
Fig~\ref{fig:scb} gives the results on the joint \textbf{Ser/Cro/Bos} dataset.
We observe a low inter-annotator agreement (\alfa\, is 0.329) and
a high variability of the classifier's performance.

\begin{figure}[!h]
\begin{center}
\includegraphics[width=\textwidth]{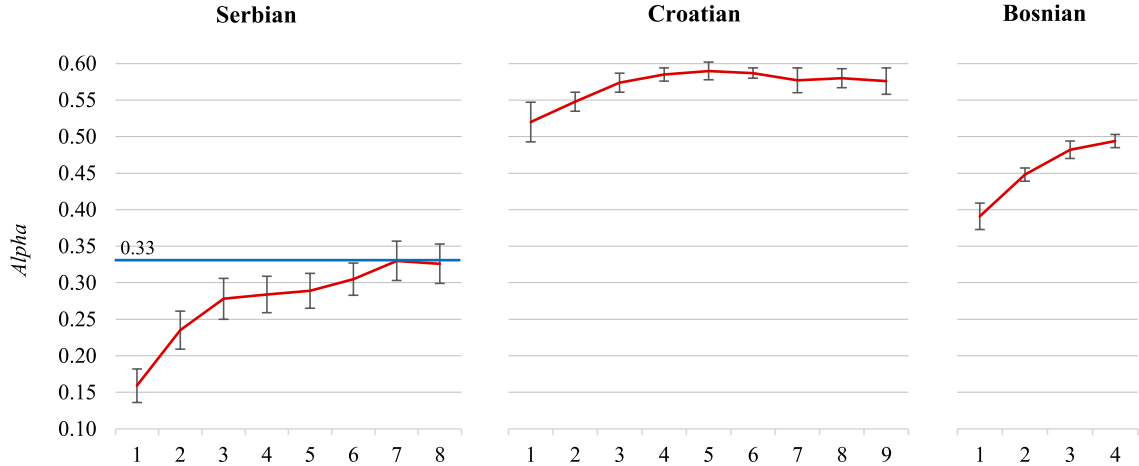}
\caption{\textbf{Separate Serbian (left), Croatian (middle), and Bosnian (right) datasets.}
Here, the lower quality Serbian set has no adverse effects on the higher quality
Croatian and Bosnian sets.}
\label{fig:ser-cro-bos}
\end{center}
\end{figure}

The three languages, Serbian, Croatian, and Bosnian, are very similar and
difficult to distinguish in short Twitter posts.
However, we argue that the reason for poor performance is not in mixing the three
languages, but in different annotation quality.
\textbf{Serbian} (73,783 tweets) was annotated by 11 annotators, 
where two of them account for over 40\% of the annotations.
All the inter-annotator agreement measures come from the Serbian only
(1,880 tweets annotated twice by different annotators, \alfa\, is 0.329),
and there are very few tweets annotated twice by the same annotator
(182 tweets only, \alfa\, for the self-agreement is 0.205).
In contrast, all the Croatian and Bosnian tweets were annotated
by a single annotator, and we have reliable self-agreement estimates.
There are 97,291 \textbf{Croatian} tweets, 13,290 annotated twice,
and the self-agreement \alfa\, is 0.781.
There are 44,583 \textbf{Bosnian} tweets, 6,519 annotated twice,
and the self-agreement \alfa\, is 0.722.
We can conclude that the annotation quality of the Croatian and Bosnian
tweets is considerably higher than of the Serbian.
If we construct separate sentiment classifiers for each language
we observe very different performance (see Fig~\ref{fig:ser-cro-bos}).
The Serbian classifier reaches the inter-annotator agreement
(albeit low) at 70,000 tweets. The Croatian classifier has much higher
performance, and reaches it maximum at 50,000 tweets (\alfa\, is 0.590).
The performance of the Bosnian classifier is also higher,
and is still increasing at 40,000 tweets (\alfa\, is 0.494).
The individual classifiers are ``well-behaved'' in contrast to the
joint Ser/Cro/Bos model in Fig~\ref{fig:scb}.
In retrospect, we can conclude that datasets with no overlapping
annotations and different annotation quality are better
not merged.

\begin{figure}[!h]
\begin{center}
\includegraphics[width=\textwidth]{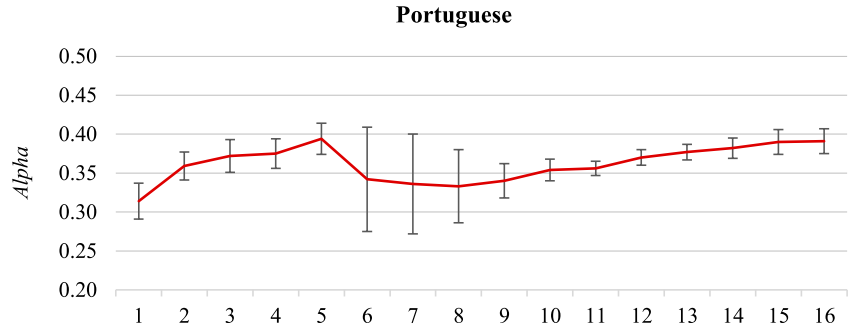}
\caption{\textbf{The Portuguese dataset.}
There are two peaks (at 50,000 tweets, \alfa\, = 0.394, and at 160,000 tweets,
\alfa\, = 0.391), and a large drop in between, due to a topic shift.}
\label{fig:por}
\end{center}
\end{figure}

\paragraph{Topic shift.}
There is no inter-annotator agreement for the \textbf{Portuguese} dataset
because only one annotator was engaged.
However, the classifier shows interesting performance 
variability (Fig~\ref{fig:por}).
After an initial peak is reached at 50,000 tweets (\alfa\, is 0.394),
there is a considerable drop and a very high variability of performance.
Inspection of the tweets (the set of 10,000 tweets added to
the first 50,000 tweets at stage 6) revealed that at the
beginning of November 2013, the Portuguese government approved
additional austerity
measures, affecting mainly public sector, to avoid the
second international bailout.
This provoked a flood of negative reactions on social media, in particular
on Twitter, and a considerable shift of focus and sentiment of Twitter discussions.
The classification model could not react immediately to the topic shift,
and it took additional 100,000 tweets to accommodate the new topics,
and the model to approach the peak performance (\alfa\, is 0.391 for
the complete dataset).

\begin{figure}[!h]
\begin{center}
\includegraphics[width=\textwidth]{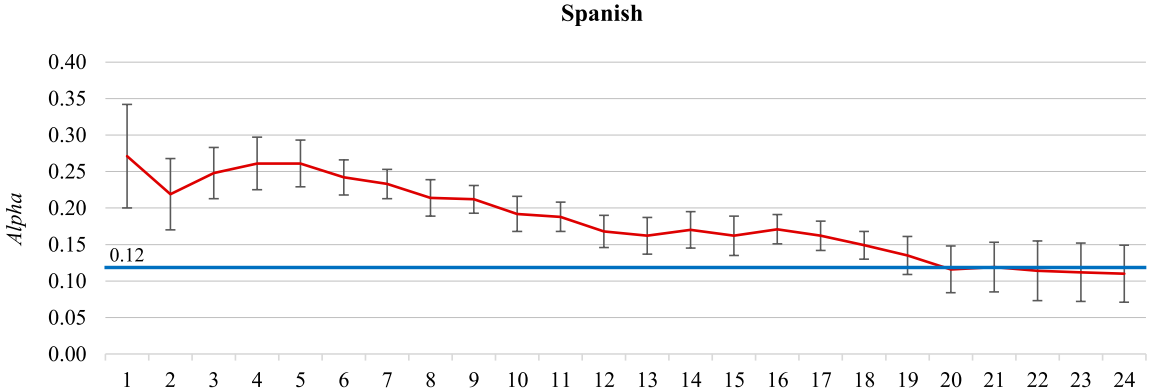}
\caption{\textbf{The Spanish dataset.}
There is a consistent drop of performance and high variability.}
\label{fig:spa}
\end{center}
\end{figure}

\paragraph{Very low annotation quality.}
What happens with the classifier's performance when the annotation
quality is low? Fig~\ref{fig:spa} shows the evolution of performance 
for the \textbf{Spanish} dataset.
We observe high variability and consistent drop in performance.
Most (over 95\%) of the Spanish tweets were annotated by one annotator,
and out of them, 40,116 tweets were annotated twice.
Therefore we have a reliable estimate of the low quality of her/his
annotations since the self-agreement \alfa\, is only 0.244.
2,194 tweets were annotated twice by two annotators and, not surprisingly,
the inter-annotator agreement is ever lower, \alfa\, is 0.120.

We observe a similar performance drop for the \textbf{Albanian} dataset
(not shown).
The main annotator (who annotated over 22\% of the Albanian tweets) has self-agreement 
\alfa\, only 0.269 (computed from 1,963 tweets annotated twice).
The inter-annotator agreement \alfa\, is only 0.126. 

Such poorly labeled data is useless for training sentiment classifiers.
However, the lesson learned is that the annotators should be monitored
throughout the annotation process, that the low quality annotators
(identified by a low self-agreement) should be excluded, and that the 
low inter-annotator agreements should be promptly investigated.

\subsection{Application datasets analyses}

The purpose of building sentiment classification models is to apply them 
in particular domains, e.g., to monitor elections or to predict stock prices.
The models are build from labeled data (where the sentiment is given) and
applied to unlabeled data (where the sentiment is to be predicted).
The models are also evaluated on the labeled data (typically by 10-fold
cross-validation) and the estimated performance can be extended to the
application if the labeled data is representative, i.e., drawn from the
same distribution as the application data. In the context of Twitter
sentiment classification this means that the labeled tweets have to be
not only language-, but also domain-specific.

In the previous subsection we analyzed the classifiers performance on
the labeled datasets and in relation to the annotator agreements.
The potential improvements can be achieved by providing additional
training data, by improving the inter-annotator agreements, and by
excluding low quality annotators. In this subsection we also consider
the relation between the training and application dataset distributions.

There are four applications where we already applied and published
Twitter sentiment classification to different domains.
Details about the sizes and distributions of the labeled and 
application datasets are in the Datasets subsection in Methods.
Sentiment distribution is captured by the \textit{sentiment score}
which is computed as the mean of a discrete probability 
distribution---details are in \cite{Kralj2015emojis}.
Here we briefly analyze and suggest possible improvements with
reference to the results in Fig~\ref{fig:agree}.

\paragraph{Facebook(it) \cite{Zollo2015facebook}.}
This is the only domain that is not limited to Twitter,
but where the same sentiment classification methodology was applied to
Facebook comments, in Italian.
There was over 1 million Facebook comments collected, and a sample of 
about 20,000 was labeled for sentiment.
The sentiment distribution in both sets is similar.
The self-agreement and inter-annotator agreement are both high,
however, there is a gap between the inter-annotator agreement
(\alfa\, is 0.673) and the classifier's performance (\alfa\, is 0.562).
Based on the lessons from the language datasets, we speculate
that 20,000 training examples is not enough, and that additional
Facebook comments have to be labeled to approach the inter-annotator
agreement.

\paragraph{DJIA30 \cite{Ranco2015eventstudy}.}
This domain deals with English tweets, but very specific for
financial markets. The sentiment labeling requires considerable domain
knowledge about specific financial terminology.
There were over 1.5 million tweets about the Dow Jones stocks collected,
and a sample of about 100,000 was annotated for sentiment.
The sentiment distribution in both sets is very similar.
The annotators self-agreement is high, but the inter-annotator
agreement is relatively low (\alfa\, is 0.438), and the classifier
even slightly exceeds it.
Also, in the period from June 2013 to September 2014, a relatively
small fraction of tweets was annotated twice (5,934), so the agreement
estimates are less reliable.
These considerations were taken into account in the subsequent period: 
from June 2014 to May 2015 altogether 19,720 tweets were annotated twice,
and the inter-annotator agreement improved for 10\% points (new \alfa\, is 0.482).

\paragraph{Environment \cite{Sluban2015sentlean}.}
This domain deals with sentiment leaning towards various environmental
issues (like climate change, fossil fuels, fracking, etc.)---not so 
well-defined problem. Consequently, the self-agreement and inter-annotator
agreement are relatively low in comparison to the Facebook(it) dataset.
Still, there is a gap between the inter-annotator agreement (\alfa\, is 0.510)
and the classifier's performance (\alfa\, is 0.397).
The training set consists of only about 20,000 labeled tweets,
and in analogy to the language datasets and Facebook(it) we conclude
that additional tweets have to be labeled to improve the classifier performance.

\begin{figure}[!h]
\begin{center}
\includegraphics[width=10cm]{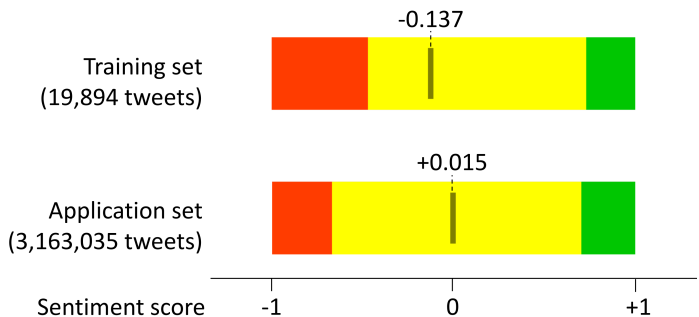}
\caption{\textbf{The sentiment distribution of the environmental tweets in
the training and application sets}. Negative tweets are denoted by red,
neutral by yellow, and positive by green color. The grey bar denotes the
sentiment score (the mean) of each dataset.}
\label{fig:env-train-appl}
\end{center}
\end{figure}

However, there is another issue. There were altogether over 3 million
tweets collected, and sentiment distribution in the training set is
considerably different from the application set (sentiment scores are
$-0.137$ and $+0.015$, respectively; see Fig~\ref{fig:env-train-appl}).
The sampling was done just in the initial phases of the Twitter
acquisition and is not representative of the whole application dataset.

We conducted an additional experiment to demonstrate the effects of different 
training and application sets.
We applied the general English language sentiment 
classification model from the previous subsection, trained on all
90,000 English tweets, to the labeled environmental tweets. 
The classifier's performance
(\alfa\, is 0.243) is considerably lower in comparison to the
environment-specific model (\alfa\, is 0.397) which was trained on
only 20,000 domain-specific tweets. The same holds for the \favg\, measure.
Detailed evaluation results
are in the Classification models performance subsection in Methods.
This result confirms our thesis that Twitter sentiment classification
is sensitive to domain of application and that sentiment labeling
has to be domain-specific.

Note also that the general English classifier has higher
accuracy (\acc\, is 0.604) than the environment-specific model
(\acc\, is 0.556). Our conclusion is that this is a clear indication
that accuracy is a misleading evaluation measure for the 
ordered three-class sentiment classification problem.

\paragraph{Emojis \cite{Kralj2015emojis}.}
There is no automated sentiment classification with the Emojis dataset.
From the 13 language datasets which consist in total of over 1.6 labeled
tweets, we selected only the tweets that contain emojis, about 70,000 in total.
The goal was to attribute the sentiment to emojis, based on the
sentiment of all the tweets in which they occur.
Fig~\ref{fig:agree} shows that Emojis is the only dataset where
the self-agreement (\alfa\, is 0.544) is lower than the
inter-annotator agreement (\alfa\, is 0.597). The reason for this anomaly is a large share
of Spanish tweets with emojis (about 20,000) that have very low
self-agreement (\alfa\, is 0.245). If we remove them from the Emojis set, 
the self-agreement increases considerably (new \alfa\, is 0.720), while
the inter-annotators agreement remains almost unchanged (new \alfa\, is 0.598).
This reconfirms our conclusion that low quality annotators 
have to be excluded and their annotations removed from the datasets.

\section{Conclusions}

We present an analysis of over 1.6 million sentiment annotated 
Twitter posts, by far the largest set made publicly available until now.
The labeled datasets are used to train sentiment classification models,
and our analysis focuses on four main aspects: quality, quantity and
sampling of the training data, and performance of the classifiers.
Our main conclusion is that the choice of a particular classifier type
is not so important, but that the training data has a major impact
on the results.

There are several specific research questions we address:

(1) What is the nature and proper formalization of the sentiment classification 
problem, in particular, are the sentiment values ordered or not?
We show that there is strong evidence that the sentiment values,
\textit{negative}, \textit{neutral}, and \textit{positive}, are
perceived as ordered by human annotators
(see subsection on Ordering of sentiment values in Methods).

(2) Which evaluation measures should be used to properly quantify 
the data quality and classifiers performance?
In all the experiment, we compute values for four evaluation measures
(\accone, \acc, \favg, and \alfa). Since there is evidence that sentiment
values are ordered, \alfa\, and \favg\, are the most appropriate as
they take the ordering into account.

(3) How to estimate the quality of the training data?
We propose to invest an extra effort to label a portion of tweets twice,
and then to compute the annotator self-agreement and the inter-annotator
agreement. The self-agreement yields a useful indication when to
exclude low quality annotators, and the inter-annotator agreement
approximates an upper bound on the performance of sentiment classifiers.

(4) How to select the most appropriate classifier?
Our results show that there are no statistically significant
differences between the top classifiers. As a consequence, one should
better direct the efforts into higher training data quality.

(5) What are acceptable levels of annotators agreement?
On the basis of the 17 datasets analyzed, we propose the following
rule-of-thumb: for self-agreement, \alfa\,$> 0.6$, and for the
inter-annotator agreement, \alfa\,$> 0.4$.

(6) How many posts should be labeled with sentiment for training?
We cannot provide conclusive answers here.
It seems that 20,000 high-quality annotations already provide
reasonable performance.
The peak performance depends on the inter-annotator agreement
and we estimate that around 100,000 annotations are needed. 
However, more important than sheer quantity is the quality, and
domain- and topic-specific coverage of the posts, as demonstrated
on several use-cases.

This gives the following directions for the short-term future work.
The annotation process has to be redesigned to allow for systematic
monitoring of the annotation quality. In particular, more than one
annotator per language/domain has to be engaged. We propose an 
increased overhead of posts to be labeled twice, from 15\% to 20\%, 
both by individuals as well as by two different annotators.
The posts to be labeled multiple times could be based on their
``importance'' as measured by their retweet 
count \cite{Sluban2015sentlean}, for example. 
The self- and the inter-annotator agreements have to be continuously
monitored and warnings issued when they drop below the selected thresholds. 
Extreme disagreements (as measured by \accone) should be promptly
directed to a ``master'' annotator who has to resolve the disagreement
and issue a proper annotation together with a brief guideline.
After each batch of a few thousand annotations, a classification
model should be trained and its performance evaluated.
This would help in monitoring the progress towards the inter-annotator
agreement as well as in detecting possible abrupt topic shifts.

There is a number of open research questions to be addressed.
One is how to combine the lexicon-based and machine learning approaches
to sentiment classification. 
In \cite{kolchyna2015twitter}, authors already showed that the combination 
of both outperforms the individual approaches.
However, sentiment lexicons are rarely available for languages
other than English and require considerable efforts to construct.
For several languages, one could use the data published
by Dodds et al.~\cite{Dodds2015lang}.
For the languages covered in this study, one can construct a
basic sentiment lexicon from the annotated tweets, in the analogy
to derivation of the emoji sentiment lexicon \cite{Kralj2015emojis}.

Another research direction, with the potential of considerable performance
improvements, is the construction and selection of 
informative features from short Twitter posts.
In this study we apply a number of standard text pre-processing
steps to extract just the textual features and eliminate noise in tweets.
However, there is a lot of additional information on Twitter to be exploited.
For example, the importance of tweets (estimated by the retweet 
count, for example),
the influence and reliability of Twitter users (estimated by their
followers, retweets, and correlations to the real-world events),
and the network features (e.g., neighbourhood and centrality)
that can be attributed to the users, and indirectly to their tweets. 
We expect that proper considerations of the broader context in which
the tweets are posted can provide for a major leap in quality and
predictive potential of the Twitter sentiment classifiers.

Finally, since the analysis of opinions expressed in social media 
is an active and evolving research area, we plan to keep up with the newest trends, 
such as performing entity-based sentiment analysis \cite{saif2013evaluation},
applying deep learning techniques \cite{dos2014deep,dos2014think,tang2014coooolll}, 
analyzing figurative language (e.g., irony or sarcasm) \cite{ghosh2015semeval},
and detecting different types of emotions (e.g., joy, sadness or anger) \cite{mohammad2015}.
The most interesting direction seems to be a shift from the basic sentiment categories
(negative, neutral, and positive) of the whole tweet, to the finer-grained emotions
about a discussed entity or topic.

\section{Methods}

\subsection*{Ethics statement}

The tweets were collected through the public Twitter API and
are subject to the Twitter terms and conditions.
The human annotators were engaged for the purpose of sentiment labeling, 
and were aware that their annotations will be used to construct the
sentiment classification models, and to estimate the
annotator self-agreement and the inter-annotator agreement.

\subsection{Datasets}

In this study we analyze two corpora of data (see Table~\ref{tab:train-num}).
The first corpus is a collection of tweets,
in 13 European languages, posted between April 2013 and February 2015.
The tweets, except English, were collected during a joint
project with Gama System (\url{http://www.gama-system.si}), 
using their PerceptionAnalytics platform (\url{http://www.perceptionanalytics.net}).
The tweets were acquired through Twitter Search API,
by specifying the geolocations of the largest cities.
For English tweets, we used Twitter Streaming API (a random sample of
1\% of all the public tweets), and filtered out the English posts.

\begin{table}[!h]
    \centering
    \begin{tabular}{r|rrr|r|c}
Dataset  & Negative & Neutral & Positive & Total & Time period\\
\hline
Albanian &  8,106 & 18,768 & 26,131 & 53,005 & June---Sep. 2013 \\
Bulgarian & 15,140 & 31,214 & 20,815 & 67,169 & Apr. 2013---Oct. 2014 \\
English & 26,674 & 46,972 & 29,388 & 103,034 & Sep. 2014 \\
German & 20,617 & 60,061 & 28,452 & 109,130 & Feb.---June 2014 \\
Hungarian & 10,770 & 22,359 & 35,376 & 68,505 & June---Aug. 2014 \\
Polish & 67,083 & 60,486 & 96,005 & 223,574 & July---Sep. 2014 \\
Portuguese & 58,592 & 53,820 & 44,981 & 157,393 & Oct.---Dec. 2013 \\
Russian & 34,252 & 44,044 & 29,477 & 107,773 & Sep.---Dec. 2013 \\
Ser/Cro/Bos & 64,235 & 68,631 & 82,791 & 215,657 & Oct. 2013---Aug. 2014 \\
Slovak & 18,716 & 14,917 & 36,792 & 70,425 & Sep.---Nov. 2014 \\
Slovenian & 38,975 & 60,679 & 34,281 & 133,935 & Jan. 2014---Feb. 2015 \\
Spanish & 33,978 & 107,467 & 134,143 & 275,588 & May 2013---Dec. 2014 \\
Swedish & 25,319 & 17,857 & 15,371 & 58,547 & Sep.---Oct. 2014 \\
\hline
Facebook(it) & 8,750 &  7,898 &  2,994 &  19,642 & Apr. 2011---Apr. 2014 \\
DJIA30  & 12,325 & 70,460 & 20,477 & 103,262 & June 2013---Sep. 2014 \\
Environment & 6,246 & 14,217 &  3,145 &  23,608 & Jan. 2014---Dec. 2014 \\
Emojis  & 12,156 & 19,938 & 37,579 &  69,673 & Apr. 2013---Feb. 2015 \\
\hline
    \end{tabular}
\caption{\textbf{The number and distribution of sentiment annotated posts, 
and the time period of the posts.} 
The top part of the table refers to the 13 language datasets,
and the bottom to the four application datasets.}
\label{tab:train-num}
\end{table}

83 native speakers (except for English) were engaged to manually label with sentiment 
over 1.6 million of the collected tweets. The annotation process was
supported by the Goldfinch platform (provided by Sowa Labs,
\url{http://www.sowalabs.com}),
designed specifically for sentiment annotation of short texts
(such as Twitter posts, Facebook comments, etc.). 
The annotators were instructed to label each tweet as either
\textit{negative}, \textit{neutral}, or \textit{positive},
by estimating the emotional attitude of the user who posted the tweet.
Tweets that were skipped or excluded are not considered in this study.

The second corpus of data are four application datasets, used in different
application scenarios and already published
\cite{Zollo2015facebook,Ranco2015eventstudy,Sluban2015sentlean,Kralj2015emojis}.

The datasets in Table~\ref{tab:train-num} are used to analyze the
annotator agreements, and to build the sentiment classification models.
The classification models build from three out of four application datasets
were actually applied to much larger sets of unlabeled data, to predict
the sentiment. Details are in Table~\ref{tab:appl-num}.
For each of the three application domains we also show the difference
between the application and training phase
in terms of the sentiment score (the mean of a discrete probability distribution,
see \cite{Kralj2015emojis} for details). For the Emojis dataset, no sentiment \
classification model was trained---the tweets with emojis were just extracted
from the above corpus of 13 language datasets.

\begin{table}[!h]
    \centering
    \begin{tabular}{r|rrr|r|cc}
         &          &         &          &       & \multicolumn{2}{|c}{Sentiment score} \\
Dataset  & Negative & Neutral & Positive & Total & application & training \\
\hline
Facebook(it) &  466,292 &   435,641 &  115,576 & 1,017,509 & $-0.345$ & $-0.293$ \\
DJIA30       &  139,132 & 1,141,563 &  275,075 & 1,555,770 & $+0.087$ & $+0.079$ \\
Environment  &  470,953 & 2,172,394 &  519,688 & 3,163,035 & $+0.015$ & $-0.137$ \\
\hline
    \end{tabular}
\caption{\textbf{Sentiment distributions of the application datasets as predicted
by the sentiment classifiers.}
The rightmost column shows the sentiment score (the mean) of the 
application and training datasets (the later from Table~\ref{tab:train-num}),
respectively.}
\label{tab:appl-num}
\end{table}

Table~\ref{tab:annotators} gives the details of the number of posts
annotated twice, by the same annotator or by two different annotators.

\begin{table}[!h]
    \centering
    \begin{tabular}{r|c|rr|rr}
Dataset  & Annotators & \multicolumn{2}{|c|}{Self-agreement} & \multicolumn{2}{|c}{Inter-agreement} \\
\hline
Albanian     & 13 &  3,325 &  6.3\% &  3,991 &  7.5\% \\
Bulgarian    & 18 &  2,083 &  3.1\% &  1,819 &  2.7\% \\
English      &\sn9 &  3,392 &  3.3\% & 12,214 & 11.9\% \\
German       &\sn5 &  6,643 &  6.1\% &  4,539 &  4.2\% \\
Hungarian    &\sn1 & 11,200 & 16.3\% &      0 &    0~~ \\
Polish       &\sn8 &  9,328 &  4.2\% & 22,316 & 10.0\% \\
Portuguese   &\sn1 &  5,350 &  3.4\% &      0 &    0~~ \\
Russian      &\sn1 & 14,452 & 13.4\% &      0 &    0~~ \\
Ser/Cro/Bos  & 13 & 19,991 &  9.3\% &  1,880 &  0.9\% \\
Slovak       &\sn1 & 11,655 & 16.5\% &      0 &    0~~ \\
Slovenian    &\sn7 & 14,482 & 10.8\% &  6,621 &  4.9\% \\
Spanish      &\sn5 & 40,205 & 14.6\% &  2,194 &  0.8\% \\
Swedish      &\sn1 &  7,149 & 12.2\% &      0 &    0~~ \\
\hline
Facebook(it) & 25 &    640 &  3.2\% &  3,262 & 16.6\% \\
DJIA30       & 10 &  2,094 &  2.0\% &  3,840 &  3.7\% \\
Environment  &\sn8 &    996 &  4.2\% &  2,718 & 11.5\% \\
Emojis       & 83 &  6,388 &  9.2\% &  3,547 &  5.1\% \\
\hline
    \end{tabular}
\caption{\textbf{The number of annotators, and the number and fraction of posts annotated twice.}
The self-agreement column gives the number of posts
annotated twice by the same annotator, and the inter-agreement column
the posts annotated twice by two different annotators.}
\label{tab:annotators}
\end{table}

The 13 language datasets are publicly available for further analyses.
Actually, our analysis reveales that it is better to partition the
Ser/Cro/Bos dataset into the three constituent languages, therefore we
provide the sentiment annotation data for the 15 languages.
The data is available as 15 language files, in the csv format, 
in a public language resource repository 
\textsc{clarin.si} at \url{http://hdl.handle.net/11356/1054}.
For each language and for each labeled tweet, there is the tweet ID
(as provided and required by Twitter), the sentiment label
(\textit{negative}, \textit{neutral}, or \textit{positive}),
and the annotator ID (anonymized).
From this data, one can compute the annotator agreement measures,
construct the ``gold standard'' training data, and train the classifiers
for different languages.

\subsection{Evaluation measures}

In general, the agreement can be estimated between any two methods of generating data.
One of the main ideas of this work is to use the same measures to estimate the agreement
between the human annotators as well as the agreement between the results of
automated classification and the ``gold standard''.
There are different measures of agreement, and to get robust estimates
we apply four well-known measures from the fields of inter-rater agreement
and machine learning.

\textbf{Krippendorff's Alpha-reliability} (\alfa) \cite{Krippendorff2012} is 
a generalization of several specialized agreement measures. 
It works for any number of annotators, and
is applicable to different variable types and metrics 
(e.g., nominal, ordered, interval, etc.).
\alfa\, is defined as follows:
$$
\mathit{Alpha} = 1 - \frac{D_{o}}{D_{e}} \,,
$$
where $D_{o}$ is the observed disagreement between annotators, and
$D_{e}$ is a disagreement, expected by chance.
When annotators agree perfectly, \alfa\;$=1$, and when the level of 
agreement equals the agreement by chance, \alfa\;$=0$. 
The two disagreement measures are defined as follows:
$$
D_{o} = \frac{1}{N} \sum_{c,c'} N(c,c') \cdot \delta^2(c,c') \,,
$$
$$
D_{e} = \frac{1}{N(N-1)} \sum_{c,c'} N(c) \cdot N(c') \cdot \delta^2(c,c') \,.
$$
The arguments, $N, N(c,c'), N(c)$, and $N(c')$,
refer to the frequencies in a coincidence matrix, defined below.
$\delta(c,c')$ is a difference function between the values of $c$ and $c'$,
and depends on the metric properties of the variable.
$c$ (and $c'$) is a discrete sentiment variable with three possible values:
\textit{negative} ($-$), \textit{neutral} (0), or \textit{positive} (+). 
We consider two options: either the sentiment
variable $c$ is nominal or ordered.
This gives rise to two instance of \alfa, $\kanom$ (nominal, when $c$ is unordered) and
$\kaint$ (interval, when $c$ is ordered), corresponding to two
difference functions $\delta$:
$$
\kanom: \;\;\;\; \delta(c,c') =
  \begin{cases}
    0 & \text{ iff } c = c' \\
    1 & \text{ iff } c \neq c' \;;
  \end{cases}
$$
$$
\kaint: \;\;\;\; \delta(c,c') = |c - c'| \;\;\;\; c,c'\in \{-1,0,+1\} \;.
$$
Note that in the case of the \textit{interval} difference function,
$\delta$\, assigns a disagreement of $1$ between the \textit{neutral} 
and the \textit{negative} or \textit{positive} sentiment,
and a disagreement of $2$ between the extremes, i.e., the \textit{negative} and \textit{positive} sentiment. 
The corresponding disagreements $D_{o}$ and $D_{e}$ between the extreme 
classes are then four times larger than between the neighbouring classes.

A coincidence matrix tabulates all pairable values of $c$ from two
annotators into a $k$-by-$k$ square matrix, where $k$ is the number of possible values of $c$.
In the case of sentiment annotations, we have a $3$-by-$3$ coincidence matrix.
The diagonal contains all the perfect matches, and the matrix
is symmetrical around the diagonal.
A coincidence matrix has the following general form:
$$
\begin{array}{c|ccc|c}
  &   & c' &   & \sum \\
\hline
  & . & . & . & \\
c & . & N(c,c') & . & N(c) \\
  & . & . & . & \\
\hline
\sum  &   & N(c') & & N \\
\end{array}
$$
In our case, $c$ and $c'$ range over the three possible sentiment values.
In a coincidence matrix, each labeled unit is entered twice,
once as a $(c,c')$ pair, and once as a $(c',c)$ pair.
$N(c,c')$ is the number of units labeled by the values $c$ and $c'$
by different annotators, $N(c)$ and $N(c')$ are the totals
for each value, and $N$ is the grand total.

The computed values of \alfa\, are subject to sampling variability,
determined by an unknown sampling distribution.
The sampling distribution can be approximated by bootstrapping
\cite{Mooney1993}.
In our case, we set the number of bootstrap samples to 1,000,
and estimate the 95\% confidence interval of true \alfa.

\textbf{F score} (\favg) is an instance of a well-known effectiveness 
measure in information retrieval \cite{VanRijsbergen1979}.
We use an instance specifically designed to evaluate the 3-class 
sentiment classifiers \cite{kiritchenko2014sentiment}.
\favg\, is defined as follows:
$$
\overline{F_1} = \frac{F_1(-) + F_1(+)}{2} \,.
$$
\favg\, implicitly takes into account the ordering of sentiment values,
by considering only the \textit{negative} $(-)$ and \textit{positive} $(+)$ labels. 
The middle, \textit{neutral}, label is taken into account only indirectly.
In general, $F_{1}(c)$ is a harmonic mean of
precision and recall for class $c$. In the case of a coincidence matrix,
which is symmetric, the `precision' and `recall' are equal, 
and thus $F_{1}(c)$ degenerates into:
$$
F_{1}(c) = \frac{N(c,c)}{N(c)} \,.
$$
In terms of the annotator agreement, $F_{1}(c)$ is the fraction of
equally labeled tweets out of all the tweets with label $c$.

\textbf{Accuracy} (\acc) is a common, and the simplest, measure of performance 
of the model which measures the agreement between the model and the ``gold standard''.
\acc\, is defined in terms of the observed disagreement $D_{o}$:
$$
\mathit{Acc} = 1 - D_o = \frac{1}{N} \sum_{c} N(c,c) \,.
$$
\acc\, is simply the fraction of the diagonal elements of the
coincidence matrix. Note that it does not account for the
(dis)agreement by chance, nor for the ordering of the sentiment values.

\textbf{Accuracy within 1} (\accone) is a special case of
\textit{accuracy within n} \cite{gaudette2009evaluation}.
It assumes ordered classes and extends the range of predictions considered 
correct to the $n$ neighbouring class values.
In our case, \accone\, considers as incorrect only misclassifications from 
\textit{negative} to \textit{positive} and vice-versa:
$$
\mathit{Acc}\!\pm\!1 = 1 - D_o = 1 - \frac{N(+,-) + N(-,+)}{N} \,.
$$
Note that it is easy to maximize \accone\, by simply classifying all the examples as
\textit{neutral}; then \accone\, $= 1$.

The four agreement measures are always computed from the same coincidence matrix.
In the case of the annotator agreements, the coincidence matrix is formed
from the pairs of sentiment labels assigned to a tweet by different annotators
(or the same when she/he annotated the tweet several times).
In the case of a classification model, an entry in the coincidence matrix
is a pair of labels, one from the model prediction, and the other from the
``gold standard''.
Experiments show that a typical ordering of the agreement results is:
\accone\, $>$ \acc\, $\approx$ \favg\, $>$ \alfa.

\subsection{The annotator agreements}

Table \ref{tab:agreement} gives the results of the annotator agreements
in terms of the four evaluation measures.
The self-agreement is computed from the tweets annotated twice by the
same annotator, and the inter-annotator agreement from the tweets
annotated twice by two different annotators, where possible.
The 95\% confidence intervals for \alfa\, are computed from
1,000 bootstrap samples.

\begin{table}[!h]
    \centering
    \begin{tabular}{r|cccc|cccc}
      & \multicolumn{4}{|c|}{Self-agreement} & \multicolumn{4}{|c}{Inter-agreement} \\
Dataset & \accone & \acc & \favg & \alfa & \accone & \acc & \favg & \alfa \\
\hline
\textbf{Albanian} & 0.961 & 0.578 & 0.601 & 0.447 $\!\pm\!$ 0.032 & 0.883 & 0.401 & 0.408 & 0.126 $\!\pm\!$ 0.032 \\
Bulgarian & 0.972 & 0.791 & 0.774 & 0.719 $\!\pm\!$ 0.030 & 0.942 & 0.510 & 0.496 & 0.367 $\!\pm\!$ 0.043 \\
English & 0.973 & 0.782 & 0.787 & 0.739 $\!\pm\!$ 0.021 & 0.966 & 0.675 & 0.675 & 0.613 $\!\pm\!$ 0.014 \\
German & 0.975 & 0.814 & 0.731 & 0.666 $\!\pm\!$ 0.022 & 0.974 & 0.497 & 0.418 & 0.344 $\!\pm\!$ 0.026 \\
Hungarian & 0.964 & 0.744 & 0.765 & 0.667 $\!\pm\!$ 0.014 & / & / & / & / \\
Polish & 0.944 & 0.811 & 0.837 & 0.757 $\!\pm\!$ 0.013 & 0.940 & 0.614 & 0.666 & 0.571 $\!\pm\!$ 0.010 \\
Portuguese & 0.890 & 0.680 & 0.741 & 0.609 $\!\pm\!$ 0.020 & / & / & / & / \\
Russian & 0.980 & 0.795 & 0.818 & 0.782 $\!\pm\!$ 0.009 & / & / & / & / \\
Ser/Cro/Bos & 0.968 & 0.764 & 0.815 & 0.763 $\!\pm\!$ 0.008 & 0.889 & 0.497 & 0.507 & 0.329 $\!\pm\!$ 0.042 \\
Slovak & 0.889 & 0.762 & 0.772 & 0.610 $\!\pm\!$ 0.015 & / & / & / & / \\
Slovenian & 0.975 & 0.708 & 0.726 & 0.683 $\!\pm\!$ 0.011 & 0.975 & 0.597 & 0.542 & 0.485 $\!\pm\!$ 0.020 \\
\textbf{Spanish} & 0.900 & 0.576 & 0.488 & 0.245 $\!\pm\!$ 0.010 & 0.829 & 0.451 & 0.423 & 0.121 $\!\pm\!$ 0.043 \\
Swedish & 0.943 & 0.740 & 0.762 & 0.676 $\!\pm\!$ 0.017 & / & / & / & / \\
\hline
Facebook(it) & 0.991 & 0.872 & 0.886 & 0.854 $\!\pm\!$ 0.037 & 0.972 & 0.720 & 0.733 & 0.673 $\!\pm\!$ 0.024 \\
DJIA30   & 0.991 & 0.929 & 0.847 & 0.815 $\!\pm\!$ 0.036   & 0.986 & 0.749 & 0.502 & 0.438 $\!\pm\!$ 0.035   \\
Environment & 0.992 & 0.823 & 0.731 & 0.720 $\!\pm\!$ 0.045 & 0.990 & 0.643 & 0.530 & 0.510 $\!\pm\!$ 0.029 \\
Emojis   & 0.933 & 0.698 & 0.700 & 0.544 $\!\pm\!$ 0.022 & 0.945 & 0.641 & 0.698 & 0.597 $\!\pm\!$ 0.025 \\
\hline
    \end{tabular}
\caption{\textbf{The self- and inter-annotator agreement measures.}
The 95\% confidence intervals for \alfa\, are computed by bootstrapping.
Albanian and Spanish (in bold) have very low agreement values.}
\label{tab:agreement}
\end{table}

Note that the Albanian and Spanish datasets have very low
\alfa\, agreement values.
All the results for \alfa, reported here and throughout the paper,
refer to the $\kaint$ instance, for the reasons outlined in the next subsection.

\subsection{Ordering of sentiment values}

Should the sentiment classes \textit{negative} ($-$), \textit{neutral} (0),
and \textit{positive} (+) be treated as nominal (categorical, unordered) or ordered?
One can use the agreement measures to estimate how are the three
classes perceived by the human annotators.

First, lets compare the agreements in terms of two variants of \alfa:
$\kaint$ (interval) and $\kanom$ (nominal). The difference between the two measures
is that $\kaint$ assigns four times higher cost to extreme disagreements
(between the negative and positive classes) than $\kanom$.
A measure which yields higher agreements hints at
the nature of sentiment class ordering as perceived by humans.
The results in Table~\ref{tab:class-order}, column two, show that
$\kaint$ always yields higher agreement than $\kanom$, except for
Spanish. We compute the average relative agreement gains by
ignoring the Albanian and Spanish datasets (which have poor
annotation quality), and Emojis (which are already subsumed by the 
13 language datasets). We observe that the average agreement is 
18\% higher with $\kaint$ than with $\kanom$.
This gives a strong indication that the sentiment classes are
perceived as ordered by the annotators.

\begin{table}[!h]
    \centering
    \begin{tabular}{r|c|cc}
Dataset & $\frac{\kaint - \kanom}{\kanom}$ & $\frac{\kaint(-,0)}{\kaint(-,+)}$ & $\frac{\kaint(0,+)}{\kaint(-,+)}$ \\
\hline
\textbf{Albanian}    & 0.786 & 0.759 & 0.146 \\
Bulgarian   & 0.193 & 0.781 & 0.547 \\
English     & 0.200 & 0.686 & 0.647 \\
German      & 0.174 & 0.702 & 0.513 \\
Hungarian   & 0.156 & 0.880 & 0.665 \\
Polish      & 0.263 & 0.685 & 0.592 \\
Portuguese  & 0.277 & 0.500 & 0.476 \\
Russian     & 0.134 & 0.693 & 0.854 \\
Ser/Cro/Bos & 0.225 & 0.706 & 0.635 \\
Slovak      & 0.001 & \textbf{1.258} & 0.956 \\
Slovenian   & 0.279 & 0.579 & 0.595 \\
\textbf{Spanish}     & \textbf{-0.158} & \textbf{1.440} & \textbf{1.043} \\
Swedish     & 0.129 & 0.806 & 0.788 \\
\hline
Facebook(it) & 0.190 & 0.675 & 0.715 \\
DJIA30       & 0.020 & 0.714 & 0.687 \\
Environment  & 0.271 & 0.559 & 0.440 \\
\textbf{Emojis}       & 0.234 & 0.824 & 0.586 \\
\hline
Average$^{*}$      & 0.179 & 0.730 & 0.650 \\
\hline
    \end{tabular}
\caption{\textbf{Differences between the three sentiment classes ($-, 0, +$).}
The differences are measured in terms of \alfa, for the union of self- and inter-annotator agreements.
The second column shows the relative difference between the $\kaint$ (interval)
and $\kanom$ (nominal) agreement measures.
The third and fourth columns show the distances of the negative ($-$) and positive ($+$)
class to the neutral class ($0$), respectively, 
normalized with the distance between them.
The last row is the average difference, but without the low quality Albanian and Spanish,
and the subsumed Emojis datasets (in bold).
Only the numbers in bold do not support the thesis that sentiment classes are ordered.}
\label{tab:class-order}
\end{table}

Second, we can use the agreement as a proxy to measure the
``distance'' between the sentiment classes.
Lets assume that the difficulty of distinguishing between the extreme
classes ($-$, +), as measured by \alfa, is normalized to 1.
If it is more difficult to distinguish between the neutral (0) and
each extreme ($-$ or +) then the normalized agreement will be lower than 1, otherwise
it will be greater than 1. The results in Table~\ref{tab:class-order},
columns three and four, indicate that for almost all the datasets the
normalized agreement is lower than 1. The only exceptions are Slovak and
Spanish. If we ignore the Albanian, Spanish, and Emojis
we observe the following average differences:
(i) it is 27\% ($1-0.73$) more difficult to distinguish between the \textit{negative} ($-$)
and \textit{neutral} (0) than between the \textit{negative} ($-$) and
\textit{positive} (+); and
(ii) it is 35\% ($1-0.65$) more difficult to distinguish between the \textit{positive} (+)
and \textit{neutral} (0) than between the \textit{positive} (+) and
\textit{negative} ($-$).

The above results support our hypothesis that the sentiment values
are ordered:
\textit{negative} $\prec$ \textit{neutral} $\prec$ \textit{positive}.
This has an implication on the selection of an appropriate performance measure
and a classification model.
The performance measure should take the class ordering into account,
therefore our selection of $\kaint$\, over $\kanom$\, is justified. 
In this respect, \favg\,
would also be appropriate, and it actually shows high correlation to $\kaint$.
The choice of an appropriate classification model is discussed in the next 
two subsections.

\subsection{Related sentiment classification approaches}

In this subsection we give an overview of the related work
on automated sentiment classification of Twitter posts.
We summarize the published labeled sets used for training the 
classification models, and the machine learning
methods applied for training.
Most of the related work is limited to English texts only.

To train a sentiment classifier, one needs a fairly large training 
dataset of tweets already labeled with sentiment. One can rely on a proxy, 
e.g., emoticons used in the tweets to determine the intended sentiment \cite{go09}, 
however, high quality labeling requires engagement of human annotators.

There exist several publicly available and manually labeled Twitter datasets. 
They vary in the number of examples from several hundreds to several thousands, 
but to the best of our knowledge, none exceeds 20,000 entries. 
Saif et al.~\cite{saif2013evaluation} describe eight Twitter sentiment datasets 
and also introduce a new one which contains separate sentiment labels for tweets and entities.
Rosenthal et al. \cite{rosenthal2015semeval} provide statistics for several of 
the 2013--2015 SemEval datasets. 
Haldenwang and Vornberger \cite{haldenwang2015uncertainty} present a 
publicly available collection of Twitter posts, which were labeled 
not only with the positive or negative sentiment, but also as uncertain or spam. 
Finally, several Twitter sentiment datasets are publicly available in CrowdFlower's 
``Data for Everyone'' collection.

There are several supervised machine learning algorithms suitable to train
sentiment classifiers from sentiment labeled tweets.
For example, in the SemEval-2015 competition, for the task on Sentiment Analysis
on Twitter \cite{rosenthal2015semeval}, the most often used algorithms are
Support Vector Machines (SVM), Maximum Entropy, Conditional Random Fields, 
and linear regression. In other cases, frequently used are also
Naive Bayes, k-Nearest-Neighbor, and even Decision Trees. 
In the following we cite several relevant papers, and report, where available, 
the comparison in performance between the algorithms used.

Go et al. \cite{go09} employ the keyword-based approach, Naive Bayes, Maximum Entropy, 
and SVM, and show that the best performing algorithm is Maximum Entropy. 
The authors in \cite{saif2014semantic} show that Maximum Entropy outperforms Naive Bayes. 
In contrast, the authors in \cite{parikh2009sentiment} report that Naive Bayes performs
considerably better than Maximum Entropy. 
Pak and Paroubek\cite{pak2010twitter} show that Naive Bayes outperforms the SVM and 
Conditional Random Fields algorithms. 
Asiaee et al. \cite{asiaee2012if} employ a dictionary learning approach, 
weighted SVM, k-Nearest-Neighbor, and Naive Bayes---Naive Bayes and its 
weighted variant are among the best performing algorithms. 
Saif et al. \cite{saif12} employ Naive Bayes for predicting sentiment in tweets.

Often, SVM is shown as the best performing classifier for Twitter sentiment. 
For example, \cite{barbosa2010robust} test several algorithms implemented in Weka,
and SVM performed best. 
The authors in \cite{kolchyna2015twitter} test the Naive Bayes, Decision Trees, and 
SVM algorithms, and find that the best performing algorithm is SVM. 
Preliminary results reported in \cite{kiritchenko2014sentiment} show that linear SVM yields
better performance than the Maximum Entropy classifier. 
Jiang et al. \cite{jiang2011target} employ SVM models for subjectivity and polarity 
classification of Twitter posts.
Davidov et al. \cite{davidov2010enhanced} employ k-Nearest-Neighbor.
Kouloumpis et al. \cite{kouloumpis11} employ AdaBoost.MH, and also test SVMs, 
but the performance results of SVMs are lower. 
Recently, researchers also applied deep learning for Twitter sentiment 
classification \cite{dos2014deep,dos2014think,tang2014coooolll}.

A wide range of machine learning algorithms is used, 
and apparently there is no consensus on which one to choose for the best performance.
Different studies use different datasets, focus on different use cases,
and use incompatible evaluation measures.
There are additional factors with considerable impact on the
performance, such as the natural language pre-processing of tweets,
and formation of appropriate features.
Typically, features are based on the bag-of-words presentation of tweets,
but there are many subtle choices to be made.

\subsection{Classification models performance}

As discussed in the previous subsection, 
there are many supervised machine learning
algorithms suitable for training sentiment classification models.
Variants of Support Vector Machine (SVM) \cite{Vapnik1995} are often used,
because they are well suited for large-scale text categorization tasks,
are robust on large feature spaces, and perform well.
The basic SVM is a two-class, binary classifier.
In the training phase, SVM constructs a hyperplane in a high-dimensional
vector space that separates one class from the other.
During the classification, the side of the hyperplane then determines the class.
A binary SVM can be extended into multi-class and regression classifiers
\cite{Hsu2002}.
For this study we implemented five extensions of the basic SVM; 
some of them take the sentiment class ordering explicitly into account.
All the SVM algorithms, and several others, including Naive Bayes \cite{Russell2003},
are implemented in the open-source LATINO library \cite{Grcar2015phd}
(a light-weight set of software components for building text mining applications,
available at \url{https://github.com/latinolib}).

\textbf{NeutralZoneSVM} is an extension of the basic two-class SVM and
assumes that neutral tweets are ``between'' the negative and positive tweets.
The classifier is trained just on the negative and positive tweets.
During the classification, the side of the hyperplane determines the
sentiment class (negative or positive). However, 
tweets which are ``too close'' to the hyperplane are considered neutral.
Various realizations of ``too close'' are described in 
\cite{Smailovic2014phd,Smailovic2015bulgelect}.

\textbf{TwoPlaneSVM} assumes the ordering of sentiment classes and
implements ordinal classification \cite{gaudette2009evaluation}. 
It consists of two SVM classifiers: 
One classifier is trained to separate the negative tweets from the
neutral-or-positives; the other separates the negative-or-neutrals 
from the positives.
The result is a classifier with two hyperplanes (nearly parallel for
all practical cases) which separates the vector space into three
subspaces: negative, neutral, and positive. During classification,
the distances from both hyperplanes determine the predicted class.

\textbf{TwoPlaneSVMbin} is a refinement of the TwoPlaneSVM classifier.
It partitions the space around both hyperplanes into bins,
and computes the distribution of the training examples in individual bins.
During classification, the distances from both hyperplanes
determine the appropriate bin, but the class is determined
as the majority class in the bin. Additionally, the classifier
can also provide the confidence of the predicted class.

\textbf{CascadingSVM} also consists of two SVM classifiers, but
does not assume that the classes are ordered. Instead, the first
classifier separates the neutral tweets (``objective'') from
the union of negatives and positives (``subjective''). The second
classifier in the cascade then considers only the ``subjective'' tweets
and separates the negatives from positives.

\textbf{ThreePlaneSVM} treats the three sentiment classes as nominal,
unordered. It consists of three binary classifiers in the
one-vs-one setting: the first separates negatives from neutrals, 
the second neutrals from positives, and the third negatives from
positives. The three independent classifiers partition the
vector space into eight subspaces. In analogy to the TwoPlaneSVMbin, 
the distribution of the training examples in each subspace
determines the majority class to be predicted during classification.

\textbf{NaiveBayes} is a well-know supervised machine learning algorithm,
and is included here for reference. It is a probabilistic classifier
based on the Bayes theorem, and does not assume ordering of the sentiment classes.

All the above algorithms were applied to the 13 language datasets and evaluated by 
10-fold cross-validation.
Standard 10-fold cross-validation randomly partitions
the whole labeled set into 10 equal folds. One is set apart for testing, the remaining
nine are used to train the model, and the train-test procedure is run
over all 10 folds. Cross-validation is \textit{stratified} when the
partitioning is not completely random, but each fold has roughly the same
class distribution. With time-ordered data, as is the Twitter stream, one
should also consider \textit{blocked} form of cross-validation \cite{Bergmeir2012},
where there is no randomization, and each fold is a block of consecutive tweets.
There are also other evaluation procedures suitable for time-ordered data,
different than cross-validation, like ordered sub-sampling, but this
is beyond the scope of the paper.
In this study we applied blocked, stratified, 10-fold cross-validation in all
the experiments. 

The Twitter data is first pre-processed by standard text processing
methods, i.e., tokenization, stemming/lemmatization (if available for a specific language),
unigram and bigram construction, and elimination of terms that do not appear
at least 5 times in a dataset. 
The Twitter specific pre-processing is then applied, i.e, replacing URLs, 
Twitter usernames and hashtags with common tokens, adding emoticon features 
for different types of emoticons in tweets, handling of repetitive letters, etc. 
The feature vectors are constructed by the Delta TF-IDF weighting scheme 
\cite{martineau2009delta}.

\begin{figure}[!h]
\begin{center}
\includegraphics[width=\textwidth]{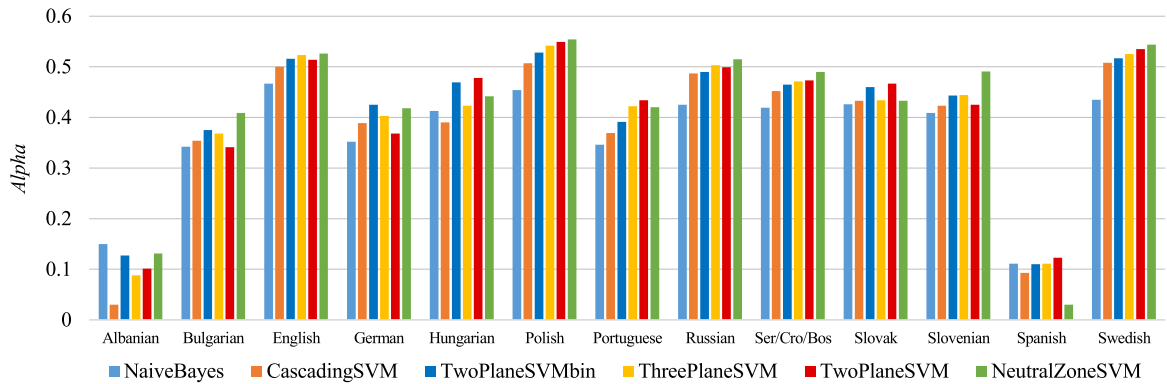}
\caption{\textbf{Comparison of six classification models in terms of \alfa.}
Results are the average of 10-fold cross-validations.} 
\label{fig:xval-alpha}
\end{center}
\end{figure}

Evaluation results, in terms of \alfa, are summarized in Fig~\ref{fig:xval-alpha}.
The classifiers are ordered by their average performance rank across 
the 13 datasets.
More detailed results, in terms of all four evaluation measures,
and also including the application datasets, are in Table~\ref{tab:xval-TwoPlaneSVMbin}.
Note that the sizes of the training datasets are lower than the numbers
of annotated tweets in Table~\ref{tab:train-num}.
Namely, tweets annotated several times are first merged into single training examples,
thus forming the ``gold standard'' for training and testing.
If all the annotations are the same, the assigned label is obvious.
If the annotations differ, the following merging rules are applied:
\textit{neutral} and \textit{negative} $\mapsto$ \textit{negative};
\textit{neutral} and \textit{positive} $\mapsto$ \textit{positive}; and
\textit{negative} and \textit{positive} $\mapsto$ \textit{neutral}.

\begin{table}[!h]
    \centering
    \begin{tabular}{r|r|cccc|l}
Dataset & Size & \accone & \acc & \favg & \alfa & Classifier \\
\hline
Albanian     &  45,758 & 0.901	& 0.460 & 0.391 & 0.127	$\!\pm\!$ 0.024 \\
Bulgarian    &  63,267 & 0.921	& 0.545	& 0.520 & 0.375	$\!\pm\!$ 0.018 \\
English      &  87,428 & 0.935	& 0.639	& 0.630 & 0.516	$\!\pm\!$ 0.006 \\
German       &  97,948 & 0.947	& 0.610 & 0.536 & 0.425	$\!\pm\!$ 0.018 \\
Hungarian    &  57,305 & 0.918	& 0.670 & 0.641 & 0.469	$\!\pm\!$ 0.012 \\
Polish       & 191,930 & 0.890  & 0.626	& 0.677 & 0.528	$\!\pm\!$ 0.011 \\
Portuguese   & 152,043 & 0.893	& 0.507	& 0.553 & 0.391	$\!\pm\!$ 0.016 & TwoPlaneSVMbin \\
Russian      &  93,321 & 0.924	& 0.603	& 0.615 & 0.490	$\!\pm\!$ 0.007 \\
Ser/Cro/Bos  & 193,827 & 0.899	& 0.559	& 0.606 & 0.465	$\!\pm\!$ 0.059 \\
Slovak       &  58,770 & 0.845	& 0.684	& 0.682 & 0.460	$\!\pm\!$ 0.013 \\
Slovenian    & 112,832 & 0.935	& 0.538	& 0.553 & 0.443	$\!\pm\!$ 0.010 \\
Spanish      & 233,204 & 0.886	& 0.531	& 0.386 & 0.110	$\!\pm\!$ 0.039 \\
Swedish      &  51,398 & 0.906	& 0.616	& 0.657 & 0.517	$\!\pm\!$ 0.022 \\
\hline
Facebook(it) &  19,642 & 0.970 & 0.648 & 0.655 & 0.562 $\!\pm\!$ 0.018 & TwoPlaneSVM \\
DJIA30       & 103,262 & 0.994 & 0.760 & 0.508 & 0.475 $\!\pm\!$ 0.013 & TwoPlaneSVM \\
Environment  &  19,894 & 0.985 & 0.556 & 0.429 & 0.397 $\!\pm\!$ 0.016 & NeutralZoneSVM \\
Environment$^{*}$ & /~~~~ & 0.965 & 0.604 & 0.344 & 0.243~~~~~~~~~~ & TwoPlaneSVMbin \\
\hline
    \end{tabular}
\caption{\textbf{Evaluation results of the sentiment classifiers.}
The dataset sizes in the second column, used for training the classifiers, 
are the result of merging multiple annotated tweets
(there was no merging for the Facebook(it) and DJIA30 datasets).
The 95\% confidence intervals for \alfa\, are estimated from 10-fold cross-validations.
The last row is an evaluation of the general English language model
(trained from the English dataset in row 3) on the Environment dataset.}
\label{tab:xval-TwoPlaneSVMbin}
\end{table}

\subsection{The Friedman-Nemenyi test}

Are there significant differences between the six classifiers,
in terms of their performance?
The results depend on the evaluation measure used, but generally
the top classifiers are not distinguishable.

A standard statistical method for testing the significant differences
between multiple classifiers \cite{Demsar2006} is the well-known ANOVA and 
its non-parametric counterpart, the Friedman test \cite{Friedman1940}.
The Friedman test ranks the classifiers for each dataset
separately.
The best performing classifier is assigned rank 1, the second best rank 2, etc. 
When there are ties, average ranks are assigned.
The Friedman test then compares the average ranks of the classifiers.
The null hypothesis is that
all the classifiers are equivalent and so their ranks should be equal. 
If the null hypothesis is rejected, one proceeds with a post-hoc test.

If one wants to compare a control classifier to other classifiers, 
the Bonferroni-Dunn post-hoc test is used.
In our case, however, all the classifiers are compared to each other,
and the weaker Nemenyi test \cite{Nemenyi1963} is used.
The Nemenyi test computes the critical distance between any 
pair of classifiers.
The performance of the two classifiers is significantly different 
if the corresponding average ranks differ by at least the critical distance.

\begin{figure}[!h]
\begin{center}
\includegraphics[width=\textwidth]{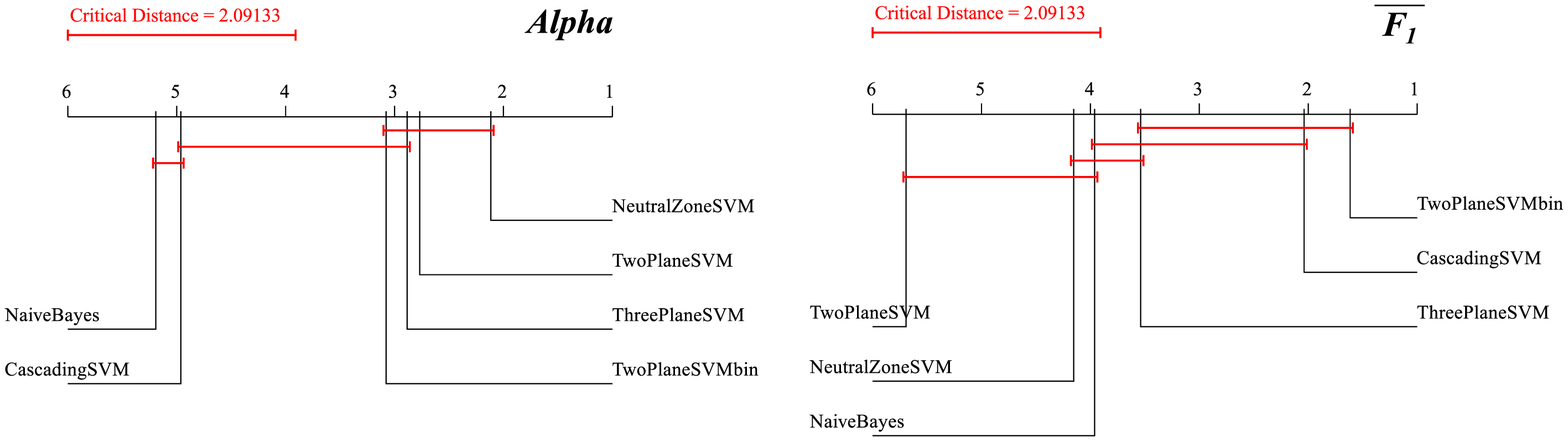}
\caption{\textbf{Results of the Friedman-Nemenyi test of classifiers ranking.}
The six classifiers are compared in terms of their ranking using two
evaluation measures, \alfa\, (left) and \favg\, (right).
The ranks of classifiers within the critical distance (2.09) are not
statistically significantly different.}
\label{fig:test-Friedman}
\end{center}
\end{figure}

Fig~\ref{fig:test-Friedman} gives the results of the Friedman-Nemenyi
test for the six classifiers trained in this study.
We focus on two evaluation measures that take the ordering of sentiment classes
into account: \alfa\, and \favg.
There are two classifiers which are in the group of top indistinguishable 
classifiers in both cases: ThreePlaneSVM (ranked 3rd) and TwoPlaneSVMbin
(ranked 4th and 1st). We decided to interpret and discuss all the
results in this paper using the TwoPlaneSVMbin classifier, since it
is explicitly designed for ordered classes.

\section*{Acknowledgments}

This work was supported in part by the European Union projects SIMPOL 
(no. 610704), MULTIPLEX (no. 317532) and DOLFINS (no. 640772), 
and by the Slovenian ARRS programme Knowledge Technologies (no. P2-103).

We acknowledge Gama System (\url{http://www.gama-system.si})
who collected most of the tweets (except English), and
Sowa Labs (\url{http://www.sowalabs.com}) for providing 
the Goldfinch platform for sentiment annotations.
Special thanks go to Sa\v{s}o Rutar who implemented several
classification algorithms and evaluation procedures in the LATINO
library for text mining (\url{https://github.com/latinolib}).
We thank Mojca Mikac for computing the Krippendorff's \alfa\, confidence intervals,
and Dragi Kocev for help with the Friedman-Nemenyi test.


\nolinenumbers

%
%


\end{document}